# Identifying charge density and dielectric environment of graphene using Raman spectroscopy and deep learning

Zhuofa Chen, Yousif Khaireddin, Anna K Swan

*Department of Electrical and Computer Engineering, Boston University, Boston*

E-mail: zfchen@bu.edu

**Abstract**

The impact of the environment on graphene's properties such as strain, charge density, and dielectric environment can be evaluated by Raman spectroscopy. These environmental interactions are not trivial to determine, since they affect the spectra in overlapping ways. Data preprocessing such as background subtraction and peak fitting is typically used. Moreover, collected spectroscopic data vary due to different experimental setups and environments. Such variations, artifacts, and environmental differences pose a challenge in accurate spectral analysis. In this work, we developed a deep learning model to overcome the effects of such variations and classify graphene Raman spectra according to different charge densities and dielectric environments. We consider two approaches: deep learning models and machine learning algorithms to classify spectra with slightly different charge density or dielectric environment. These two approaches show similar success rates for high Signal-to-Noise data. However, deep learning models are less sensitive to noise. To improve the accuracy and generalization of all models, we use data augmentation through additive noise and peak shifting. We demonstrated the spectra classification with 99% accuracy using a convolutional neural net (CNN) model. The CNN model is able to classify Raman spectra of graphene with different charge doping levels and even subtle variation in the spectra between graphene on $SiO_2$ and graphene on silanized $SiO_2$. Our approach has the potential for fast and reliable estimation of graphene doping levels and dielectric environments. The proposed model paves the way for achieving efficient analytical tools to evaluate the properties of graphene.

**Introduction.**

Raman spectroscopy is a versatile tool using vibration information of atoms to reveal the fingerprint of materials[1]. Traditional analysis methods of Raman spectra require data preprocessing and peak fitting[2] to extract parameters that are used in estimating the properties of materials. For instance, peak parameters such as peak position and peak width are extracted from graphene Raman spectra to estimate strain and charge doping levels[3–5]. The preprocessing might include manual calibration peak shift, cosmic ray removal, baseline correction, and spectral smoothing[6,7], all of which could introduce artifacts and variability. The analysis requires peak fitting and peak parameter extraction such as peak position, intensity, and width. These parameters may not suffice to reveal all the material's properties. Peak asymmetry and inter-relational spectral properties may also be necessary[5]. Additionally, collected spectroscopic data vary slightly due to different experimental setups and environments. Even using the same setup, the collected spectra measured with different focal planes and signal-to-noise ratio (SNR) lead to different peak information if analyzed using traditional methods. These variations, artifacts, and environmental differences pose a challenge in analysis.

In recent years, the development of machine learning and powerful deep learning models have offered exciting tools in material science, especially spectral data analysis[8]. Machine learning algorithms can extract features and learn correlations between features within the spectrum. Various machine learning algorithms have been previously applied to Raman spectra analysis. For instance, principal component analysis (PCA) was used to reduce the dimensionality of Raman spectra and analyze food colorants[9]. Support vector machines (SVMs) were adopted in classifying Raman spectra of foreign fats and oils discrimination[10]. Multivariate classification algorithms such as K-nearest neighbors (KNN) and Partial Least Squared (PLS) discriminant analysis were applied to olive and vegetable oils classification[11]. Random forest (RF) was used to classify Raman spectra for dengue fever analysis of infected human[12], etc. While these machine learning algorithms show high prediction accuracies, most of them suffer from poor generalization and computing inefficiencies which limit their applications[13–16].

Deep learning, a class of machine learning algorithms, uses artificial neural networks to progressively learn and extract features from input data. These networks then use the high-level extracted features to classify or process the input data, which offers a better generalization comparing to processing data using machine learning algorithms. Deep learning offers unprecedented opportunities in feature extraction when it comes to datasets such as Raman spectra. They are capable of efficiently handling large amounts of data and detecting complex nonlinear relationships[17]. With the development of deep learning, more deep neural networks have been proposed and applied to Raman spectral analysis[18–23].

Graphene has applications in sensors, displays, and flexible devices due to its superior transport properties, e.g. giant intrinsic mobility and distinctive electronic structure[24–26]. Estimating the charge density is important to evaluate the device quality of these applications, and the dielectric environment of graphene affects the phonon signatures in graphene, so it should be known[27–30]. The benchmark method to determine the quality of a device is the electrical transport measurement. The mobility, the residual charge doping level, and the overall charge density variations can be determined with high precision. However, transport measurements require a labor-intensive fabrication process that may in turn alter the mobility of the material. Raman spectroscopy is a non-invasive tool that reveals the information of charge density and dielectric screening effect in graphene[1,30]. However, the subtle variations in graphene's Raman spectra due to charge variation or dielectric change is difficult to determine in detail using traditional data preprocessing and curve fitting. For instance, the Raman spectrum peak position and peak width are similar for graphene on $SiO_2$ and graphene on silanized $SiO_2$ surface. Also at low doping levels, it is difficult to manually differentiate Raman spectra of graphene with different low charge density (< $2\times 10^{12}$ $cm^{-2}$)[4]. Therefore, a more efficient and accurate method to analyze graphene's Raman spectra is extremely desirable. Given the power of deep neural networks, we aim to develop an architecture to accurately and efficiently differentiate graphene charge densities and dielectric environments.

In this work, we analyze Raman data collected from several large areas (spatial Raman maps) from graphene in varying environments. In addition, we explicitly consider experimental flaws e.g., drift in the focal positions for a long acquisition time, or from low signal-to-noise ratio (SNR), where noise is added to low noise spectra (Figure S1-3). We develop multiple deep learning models and machine learning algorithms to classify graphene Raman spectra based on different charge densities and dielectric environments. A schematic of Raman spectra collections and a deep learning model classification is shown in Figure 1. For Raman spectra with high signal to noise ratio (SNR), both deep learning models and machine learning algorithms show high classification accuracy (95% to 99%). The deep learning models show better generalization and are less sensitive to noise. The proposed CNN can successfully identify the subtle variations caused by the internal charge variation and the external dielectric differences. The models trained here can efficiently and automatically estimate the charge density of graphene and identify its dielectric environment to aid with the phonon property analysis of Raman spectroscopy.

## Results and discussion

Two datasets are used in this work, the graphene charge variation dataset (charge dataset) and the graphene dielectric variation dataset (dielectric dataset). Graphene samples with different charge densities are prepared by exfoliating graphene on silane-treated and untreated $SiO_2$ surfaces. Different charge doping levels are controlled by the surface roughness of the silane layer (Octadecyltrimethoxysilane, OTMS). The charge dataset is collected by performing Raman spatial mapping on each graphene sample with different charge doping levels. Even though treating substrates with OTMS yields a big difference in graphene charge density as measured by transport data[31], the line shape of corresponding Raman spectra changes minimally. The dielectric dataset is collected on graphene samples in different dielectric environments, which are formed through the dry transfer method[32] with exfoliated boron nitride (BN). The four different dielectric environments are: graphene/$SiO_2$, graphene/OTMS/$SiO_2$, BN/graphene/$SiO_2$, and BN/graphene/BN. To analyze the collected Raman spectra data, we try to use deep learning and machine learning techniques to classify these Raman spectra as shown in Figure 1a, b. Most of the collected data in our set up has a low noise level, < 1%. Due to different charge and screening effects, the Raman spectra of graphene on hBN and graphene on $SiO_2$ can be easily differentiated by checking the 2D peak position and G width. On the other hand, it is difficult to differentiate between the Raman spectra from graphene on $SiO_2$ and graphene on OTMS-treated $SiO_2$. Moreover, some traditional mathematic methods, e.g. principal component analysis cannot separate different Raman spectra classes, shown in Figure S7. Thus, we explore the capabilities of deep learning and ML to categorize these spectra with charge density in the range of $10^{11}$- $4\times10^{12}$ $cm^{-2}$, and spectra of graphene from different substrates including the difference between graphene on $SiO_2$ and graphene on OTMS treated $SiO_2$.

The collected Raman spectra are cropped into the same range (1450 to 3152.4 $cm^{-1}$) to display the G and 2D bands. The input data, one by 728 one-dimensional vector, are fed into machine learning or deep learning models. The output of these models is a 4 by 1 vector, corresponding to four different accidental doping levels, or four different dielectric environments. Detailed information about Raman measurement, deep network architectures, and machine learning algorithms are shown in Methods. The Raman spectra of each class in the graphene charge variation dataset (charge dataset, total 2112 Raman spectra) are color-grouped and plotted in Figure 1c. For each sample, we perform a Raman spatial mapping with an average of 528 spectra to ensure the collected spectra capture enough variation within the sample. Figure 1b also shows the block diagram of a representative one-dimensional CNN and a schematic of the classification results. The classification into 4 groups, C1-4, here corresponds to the charge doping levels of graphene in the charge dataset. For the graphene dielectric variation dataset (dielectric dataset, total 4419 Raman spectra), the classification result from the phonon signatures of graphene samples will relate to different dielectric environments.

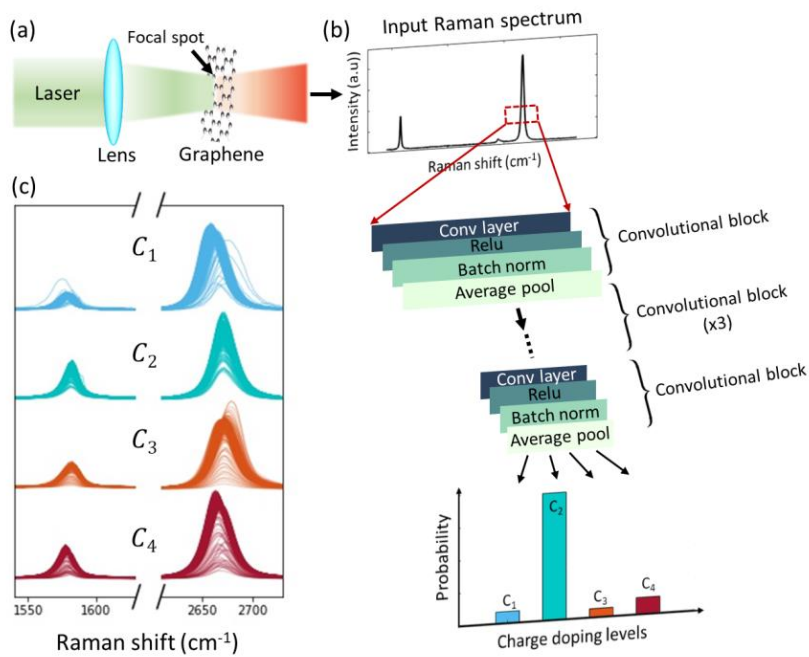

*Figure 1 Predicting the doping level of graphene using a convolutional neural network (CNN). (a) A conceptual schematic of Raman spectrum measurement on a graphene sample. (b) A representative Raman spectrum of graphene. The spectrum is fed into a one-dimensional CNN with 5 convolutional blocks and classified as one of the 4 classes (C1-4), corresponding to 4 different charge doping levels of graphene samples. (c) Raman spectra of graphene sample with 4 different charge doping levels. The peak profile can be difficult to distinguish. An average of 528 spectra from 4 different classes is shown in the (c).*

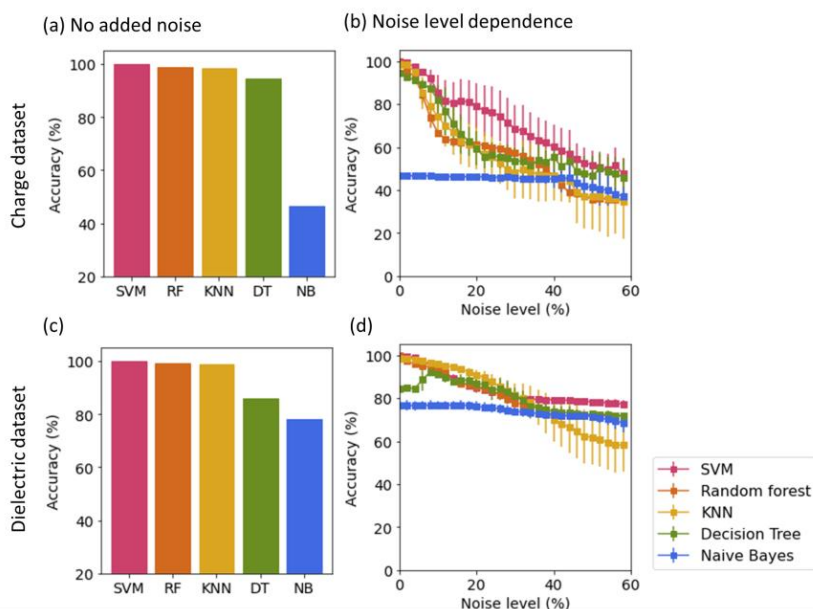

*Figure 2 Performance of 5 machine learning algorithms: Support vector machine (SVM), Random Forest (RF), K-nearest neighbors (KNN), Decision Tree (DT), and Naïve Bayes (NB). (a) Testing accuracy of the machine learning algorithms on the charge dataset without noise. (b) Testing performance as a function of noise on the charge dataset. (c-d) Testing results of the machine learning algorithms on the dielectric dataset.*

**Performance of machine learning algorithms**

We implement 5 commonly used machine learning algorithms and record their performance on both datasets. The implemented machine learning algorithms include Support vector machine (SVM), Random Forest (RF), K-nearest neighbors (KNN), Decision Tree (DT), and Naïve Bayes (NB). The parameters of each of these algorithms are tuned to achieve the highest classification accuracy using the dielectric dataset, shown in Figure S8. Figure 2a and c show the prediction accuracy of the machine learning algorithms on each dataset. KNN, RF, and SVM perform very well in both datasets achieving high accuracies in the range of 97 – 99 %. Hence, although these algorithms are simpler to implement than the deep architectures, they are still able to model the complex features present in the Raman spectra well.

However, Naïve Bayes shows poor performance in both datasets with an accuracy of 46 % and 76 %, respectively. This means the feature-independent assumption in Naïve Bayes does not ring true in the collected Raman data[33]. On the other hand, Decision Tree shows a high prediction accuracy of 94 % for the charge dataset and 84 % for the dielectric dataset. It is a non-parametric algorithm that classifies a population into branches, which are interpretable and robust to outliers. Although both algorithms are similar, Random Forest is stronger at generalization than Decision Tree. This is confirmed by our experimental results, Figure 2a and c, where Random Forest achieves higher accuracy.

Signal to noise ratio plays an important role in Raman spectra analysis. To explore this, we artificially generate different noise, add them to our collected graphene Raman spectra, and fit the spectra with Voigt profile to analyze its peak information. Details about generating noise and how noise can impact the analytical results are shown in SI. Here, we define noise level as the noise fluctuation range over the maximum intensity of the Graphene Raman 2D peak. The results show that the extracted parameters are highly dependent on noise; high noise levels lead to larger parameter fluctuations. To study the effect of noise on the machine learning algorithms, we trained and tested all models at various noise levels. At each noise level, the machine learning models were run 100 times and their statistical results (average and standard deviation) are shown in Figure2b and d.

As expected, increasing the noise level inherently increases the complexity of the data and reduce the predicting accuracy of all the machine learning models. In Figure 2b, as the noise level increases, the performance of SVM, RF, KNN, and DT decreases to around 30 – 50 %. Amongst all algorithms, KNN decreases the fastest with noise. This is because KNN classifies data by calculating the distance between two data points and groups the data points using its neighbors[34]. This tends to find the most similar samples and adding noise in each spectrum may confuse spectra between different classes. A similar performance for KNN is seen on the dielectric dataset shown in Figure 2d. The prediction results of the NB are less sensitive to the noise effect while shows a low classification accuracy. In Figure 2d the machine learning models are trained on the dielectric dataset. The predicting accuracy of SVM, DT, RF, and NB converge to around 70~80 % as noise level increases, around 20 % higher than that achieved in the charge dataset. This is because changes in the dielectric environment cause a substantial shift in the 2D phonon of graphene. Hence, dielectric environment differences have a larger impact on the Raman spectra, thus making it easier for machine learning classification. We also note that the accuracy of the decision tree algorithm increases as the noise level increases from 5 % to 10 %. This may be because at such large noise levels, the model is no longer able to overfit the data due to the complexity presented by high noise, enabling it to perform better.

**Performance of deep learning models**

It is important to note that most of machine learning algorithms suffer from three things: high computational complexity, large memory utilization, and limited generalizability. The poor generalizability of machine learning algorithms makes it difficult for real application since most of the collected Raman spectra data may has a certain noise level, leading to low prediction accuracies. Thus, to find the optimal machine learning tool for graphene-based Raman spectra data analysis, we looked into deep learning approaches. We implemented 4 different deep architectures: Convolutional Neural Network (CNN), Fully Convolutional Neural Network (FullCNN), Fully Connected Neural Network (FC), and MultiHead Convolutional Neural Network (MHCNN). We then compare the performance of these models on both datasets. The parameters of each architecture are tuned to achieve the highest classification accuracy within each dataset. Detailed model architectures and training processes are described in Methods and SI. Briefly, a block of the normal CNN, in our models includes a convolutional layer, a ReLU activation layer, a batch-normalization layer, and a pooling layer. The FullCNN is constructed using only convolutional layers. The FC is built using densely connected layers. The MHCNN consists of three different CNN blocks in parallel.

To compare the performance of these deep learning models with previous machine learning algorithms, we carried out the same noise experiment discussed above. Figure 3a shows the overall testing accuracy of the 4 models trained on the charge dataset. All models implemented achieve a high testing accuracy with a minimum of 94% accuracy. The top performance, 99.8 %, is achieved by the CNN model. Of all four models, the MHCNN has the poorest performance. This can be attributed to the fact that this network is comparatively shallower and wider as compared to the deeper and narrower architectures of all 3 other models. These deeper and narrower architectures can capture more complex information in the Raman spectra. Next, comparing the FullCNN and FC, the FullCNN performs better. This can be attributed to its convolutional blocks. Their ability to share weights and scan for patterns enables allow for more effective feature extraction of the Raman spectra. Regardless, the best performing network, the CNN, is the one that combines the benefits of all the above. It is a deep network using convolutional blocks and ending with fully connected blocks. Using all the advantages of the previous networks gives it an upper edge in performance.

Figure 3b shows the confusion matrix of the CNN model on the charge dataset, showing high classification accuracy. The misclassification of Raman spectra from Graphene on $SiO_2$ might be due to the large charge fluctuation on the whole sample due to charge puddling. The misclassification of Raman spectra from Graphene/OTMS might be because of the poor coverage of the OTMS layer. There are some charge puddles with similar doping levels on both graphene samples, which can be confirmed by the overlap of charge density range obtained from transport measurement, shown in SI.

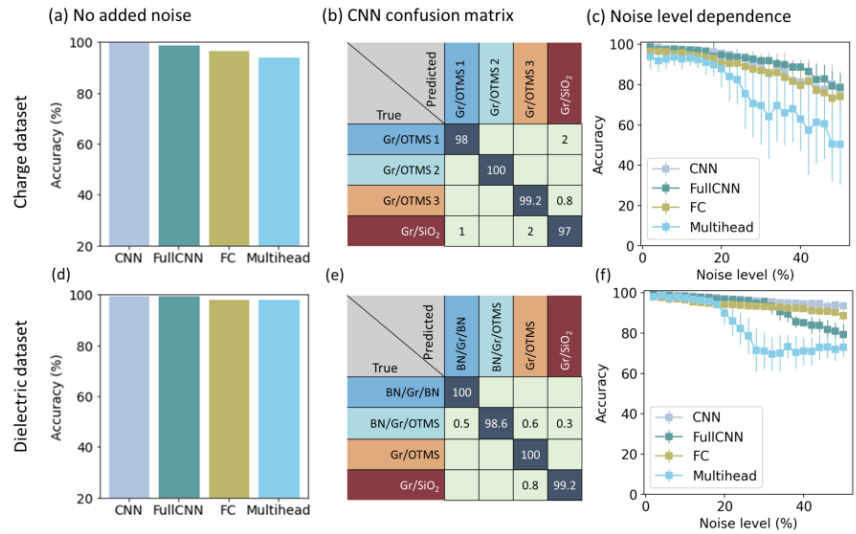

*Figure 3 Performance of 4 deep learning models: MHCNN, FCNN, FullCNN, and CNN. (a) Testing accuracy of the deep models trained on the charge dataset without noise. (b) the confusion matrix of the best model, CNN, on the charge dataset. (c) Testing performance as a function of noise, on the charge dataset. (d-f) Testing results of the deep models trained on the dielectric dataset. The error bars in (c) and (f) are calculated by running the classification process for 100 times.*

To study the effect of noise on our proposed models, we trained and tested all models at various noise levels. At each noise level, the 4 proposed deep learning models were run 100 times with random initialization and their statistical results (average and standard deviation) are shown in Figure 3c. Increasing the noise level inherently increases the complexity of the data. As expected, all models perform best with noise levels less than 10 % (high SNR) and their performance degrades as the noise level increases. At the highest tested levels of noise, the FullCNN shows the most resilience. Otherwise, the CNN and FullCNN are comparably showing strong robustness to noise and effectively capturing information from the complex dataset. On the other hand, the MHCNN's accuracy degrades from 90 % to around 60 % as the noise level increases. This can once again be attributed to its shallow and wide architecture. Figure 3d-f show similar results on the dielectric dataset. Again, the CNN outperforms other models with an overall accuracy of 99 %. Note that in Figure 3f, the CNN model performs better than the FullCNN model even at high noise levels. The MHCNN degrades even faster as the noise level increases.

As shown by the experimental results, all our proposed deep models perform much better than the commonly used machine learning algorithms which are more sensitive to noise, causing the classification accuracy to decrease rapidly with increasing levels of noise. This is important since collected Raman spectra usually have various SNRs. On the other hand, our best performing CNN achieves a high accuracy of over 99% and maintains an accuracy of over 90% up to a noise level of 30 %. This means that our model has wide applications in graphene Raman spectra analysis. For collected Raman spectra data with high SNR (or noise level < 1%), you can use both deep learning models or machine learning algorithms. It is much more desirable to use deep learning models if the collected data has a low SNR since the machine learning algorithms do not generalize well and their performance degrades heavily with noise.

**Stability and reproducibility of deep architectures**

Due to the inherent randomness present in deep learning such as random initializations, shuffling, and stochastic optimization steps, our models may converge to slightly different classification accuracies every time we train them. In this section, we next investigate if we can improve the stability and reproducibility of our proposed deep learning models using data augmentation, including peak shifting and additive noise. Peak shifting is done by randomly shifting the Raman spectra within $\pm$ 30 $cm^{-1}$, and the augmentation through noise is described in Methods. As an example, we set the noise level to 5%. The SNR of the raw data is 0.8 - 0.9 %. For each model, more than 400 experiments are run and the data is reshuffled between training and testing each time. Figure 4 shows the pie charts of classification accuracy of the different models trained on both datasets. Figure 4a and c shows the pie charts of classification accuracy for different deep learning models trained on the charge dataset and dielectric dataset, respectively. The distribution of red lines is mostly located within 90 – 100 % accuracy. However, the optimized loss function may converge to local minima yielding low accuracies ranging between 30 – 80 %. The MHCNN seems to be the most prone to converge at these local minima, and other deep learning models seems to be stable and almost consistently achieving 90 – 100 % accuracy. Figure 4b and d shows the results of all models trained on the charge dataset and dielectric dataset with data augmentation. The results show less red lines distributed between 30 to 80%, especially the Multihead model. Data augmentation improve the data generalization and decreases prediction variations in these models. Moreover, the CNN, FullCNN, and Fully Connected models are insensitive to the added noise since the distribution of prediction accuracy doesn`t have a big difference after applying data augmentation.

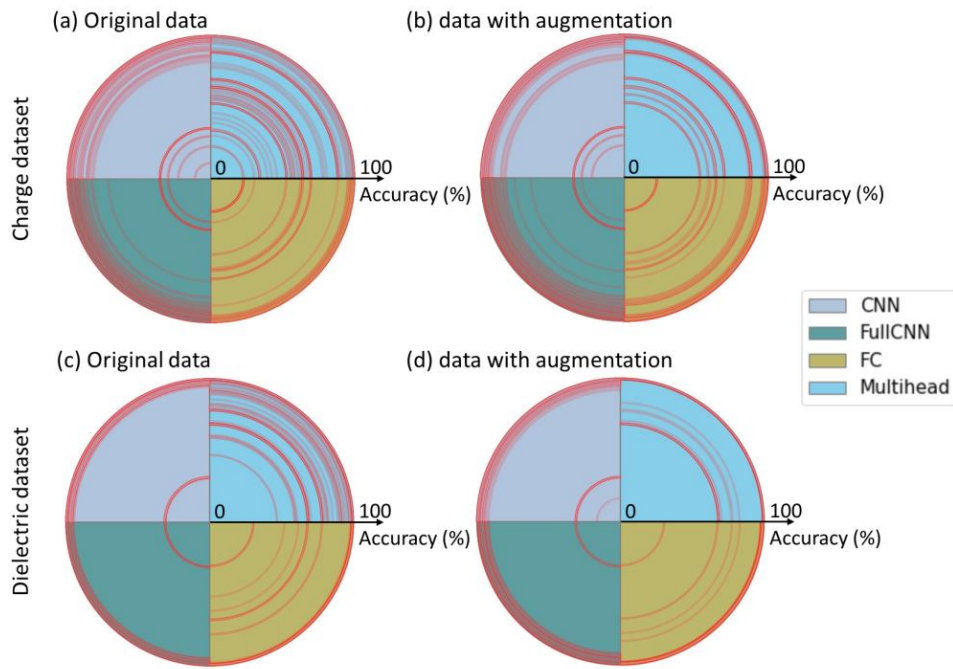

*Figure 4 Model reproducibility and performance improvement after data augmentation. (a) A pie chart of classification accuracy for different models on the charge dataset. More than 400 times of train/test data splitting and model training were performed for each model. The red lines in inner circles indicates low prediction accuracy while the red lines locate in the outer circles indicates high prediction accuracy. (b) A pie chart of classification accuracy for different models on the charge dataset with data augmentation. (c-d) pie charts of classification accuracy for different models on the dielectric dataset with/without data augmentation.*

**Conclusion**:

We have tested machine learning models and deep learning CNN models to characterize the charge density and dielectric environment of graphene from Raman spectra. The CNN models perform better than the ML models, especially for lower signal to noise data. Four different deep architectures (CNN, FullCNN, FC, MHCNN) are implemented and tested. The CNN model shows excellent performance with an accuracy 95 – 99 % on the graphene charge variation dataset and graphene dielectric variation dataset. This CNN model is able to classify Raman spectra of graphene with different charge doping levels and even subtle variation in the spectra between graphene on $SiO_2$ and graphene on silanized $SiO_2$. Data augmentation, including adding noise and shifting Raman peaks, strengthens the performance of all models on both datasets. We also compare the deep models with commonly used machine learning algorithms (SVM, RF, KNN, DT, NB) on the datasets, with and without augmentation. The results show that the CNN has excellent performance and is less sensitive to noise than the tested machine learning algorithms. Both deep learning models and machine learning algorithms can be adopted for Raman spectra classification if the collected Raman spectra has high SNR. However, deep learning models are favorable for Raman spectra with low SNR since machine learning algorithm is not generalized, and the performance is highly affected by noise levels. For specific datasets and real applications, the proposed CNN model requires some tweak to optimize the performance. With development of deep learning libraries and frameworks, setting up such models for Raman spectra analysis becomes much easier and faster. By combining Raman spectroscopy with deep learning, our approach can quickly and reliably classify graphene Raman spectra and estimate graphene doping levels and dielectric environments.

**Methods**

*Sample preparation*: 1. The graphene charge variation dataset (charge dataset) is collected by performing Raman spatial mapping on each graphene sample with different charge doping levels. Graphene samples with different charge densities are prepared by exfoliating graphene on OTMS-treated and untreated $SiO_2$ surfaces. Different charge doping levels are controlled by the surface roughness of the OTMS layer. The doping level of each sample is evaluated by transport measurement, shown Talbe1 in supplementary information (SI). C1 has doping range: 24 to $43 \times 10^{11}$ $cm^{-2}$, C2 has doping range: 16.7 to $28.1 \times 10^{11}$ $cm^{-2}$, C3 has doping range: 3.7 to $18.1 \times 10^{11}$ $cm^{-2}$, and C4 has doping range: 1.4 to $4.8 \times 10^{11}$ $cm^{-2}$. 2. The graphene dielectric variation dataset (dielectric dataset) is collected by using the same Raman spatial mapping method. Graphene samples are prepared by exfoliation and the different dielectric environments are fabricated through the dry transfer method[32] with exfoliated boron nitride (BN). Four different dielectric environments are constructed: graphene/$SiO_2$, graphene/OTMS/$SiO_2$, BN/graphene/$SiO_2$, and BN/graphene/BN. A schematic of graphene in different dielectric environment is shown in Figure S4 in SI.

*Data acquisition:* All the spectra are collected using Renishaw Raman spectroscopy with a 532nm laser using the same illumination power of 300 μW and a grating of 1200 groves/mm. The ×100 (0.7 NA) objective lens generates a diffraction-limited spot size of 0.45 μm in diameter. The Raman spatial mappings are taken with a 1 μm step size to avoid overlap between spectra. The collected spectral ranges between 662.05 and 3152.4 $cm^{-1}$. All the spectra data are cropped into a smaller range (1450 to 3152.4 $cm^{-1}$) to eliminate the effect of the Boron nitride Raman peak during the training process. The charge dataset consists of 4 different classes with the following counts: Gr/OTMS1 (484), Gr/OTMS2 (633), Gr/OTMS3 (753), and Gr/$SiO_2$ (242). The dielectric dataset also consists of 4 different classes with the following counts: Gr/$SiO_2$ (1355), Gr/OTMS/$SiO_2$ (1386), BN/Gr/OTMS (727), and BN/Gr/BN (951). Raman spectral plots of each class are shown in Figure S5.

*Data preprocessing and augmentation*: The cosmic rays are removed through MATLAB before any further processing. All the spectra are first rescaled between 0 and 1 to speed up the convergence in the training process. The data augmentation process is designed to mimic and capture the expected variations in Raman signals that may happen during different experimental setups and environments. Data augmentation includes adding different noise levels (up to 50 %) and shifting the Raman peaks (up to $\pm$ 30 wavenumber). This can easily happen during different experimental conditions such as different power, misalignment, and different focus plane in the experimental setup. This, as shown previously, improves the generalization of the models on the data sets. An illustration of data augmentation is shown in Figure S6.

*Flowchart of experimental design*: Before feeding into any models, all the raw data is rescaled and the cosmic rays are removed. In the data augmentation step, we applied peak shifting to better capture all expected variations in the collected spectra. Each model is studied with different noise levels. 5 different machine learning models (SVM, RF, KNN, DT, NB) and 4 different deep learning models (CNN, FullCNN, FC, and MHCNN) are implemented, shown in Figure 5. All the models are trained, tuned, and optimized and the optimal performances are compared.

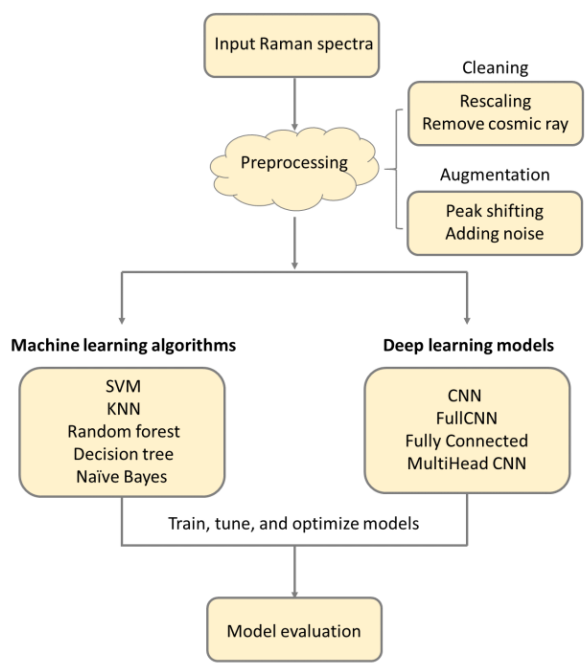

*Figure 5 Flowchart of the experimental design. After preprocessing, 5 different machine learning models (SVM, RF, KNN, DT, NB) and 4 different deep learning models (CNN, FullCNN, FC, and MHCNN) are implemented and their optimal performances are compared.*

*Deep architectures, machine learning algorithms, and training details:* To achieve high prediction accuracy, we develop 4 different neural networks including normal CNN, FullCNN, MultiHead CNN, and Fully Connected neural network (FC). The normal CNN consists of 5 blocks, each with a one-dimensional convolution layer, a rectified linear activation unit (ReLU), and a batch normalization layer. The FullCNN model consists of 5 one dimensional convolution layers with decreasing filter sizes. The MultiHead CNN consists of 3 different parallel paths each of which contains different CNN blocks. The FC is the combination of 7 back-to-back linear layers. Detailed model architectures and model summaries can be found in SI. We initially trained and tuned each model using a stratified 60/20/20 train/val/test split. Once all required parameters were tuned, we then divide the data using a stratified 80/20 train/test split in the following experiments. The models were trained using the Adam optimizer with a tuned learning rate of 0.001. The criterion used for all the models is a categorical cross-entropy. Each model performs better with a different batch size: CNN used 64, Full CNN used 16, MHCNN used 128, and the FC used 64. Model summaries of the deep learning models are shown in Table S2-5 in SI.

All models were trained for up to 100 epochs. All network architectures were built, trained, and tuned using Pytorch. The machine learning classification algorithms serve as a benchmark for comparison on both charge and dielectric datasets. The machine learning algorithms are implemented using Scikit-learn library[35] and their parameters are tuned to maximize performance on each of the above datasets. Classification accuracy is adopted as evaluation metrics since the number of spectra in each class is comparable.

The machine learning algorithms are trained and tuned using the dielectric dataset since the dielectric dataset has more variation. All models are trained using a stratified 80/20 train/test split. Some typical hyper parameters of each algorithm are trained and optimized with the highest prediction accuracy.

https://doi.org/10.1039/C6CS00915H.

# Supplementary information

Identifying charge density and dielectric environment of graphene using Raman spectroscopy and deep learning

Zhuofa Chen, Yousif Khaireddin, Anna Swan

Department of Electrical and Computer Engineering, Boston University, Boston

E-mail: zfchen@bu.edu

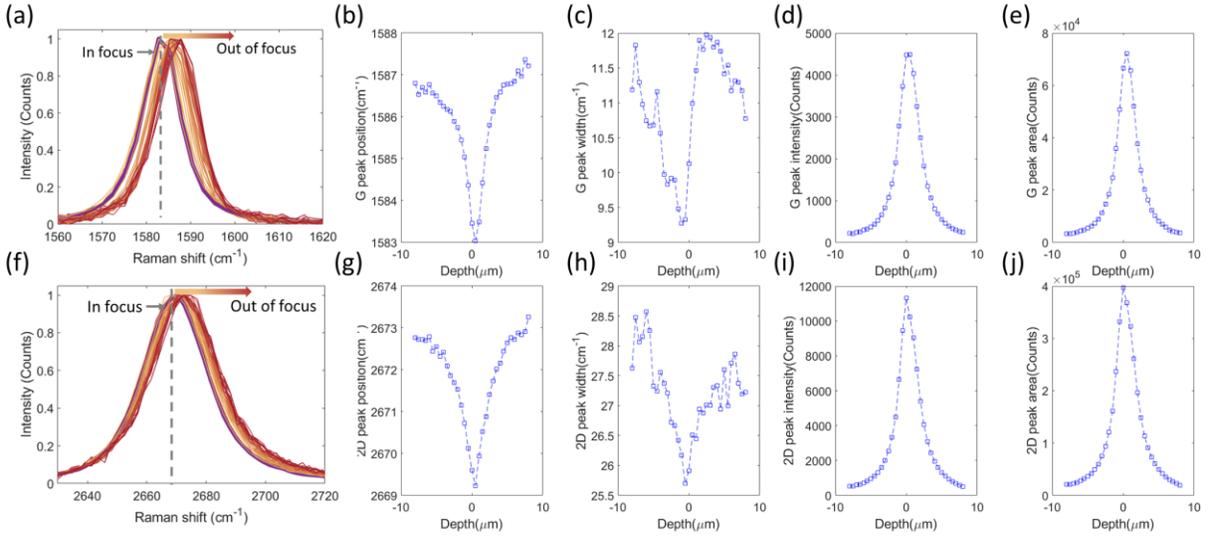

*Figure S 1 Effect of focal plane on Raman spectra and the analysis results by curve fitting. (a) Raman G peak of graphene at different focus planes. (b-e) The curve fitting results of G peak position, G peak FWHM, G peak intensity, G peak area. (f) Raman 2D peak of graphene at different focus planes. (g-j) The fitting results of 2D peak position, 2D peak FWHM, 2D peak intensity, 2D peak area. The Raman spectra have different peak information at different focus planes even though we maintain all the other experimental conditions constant.*

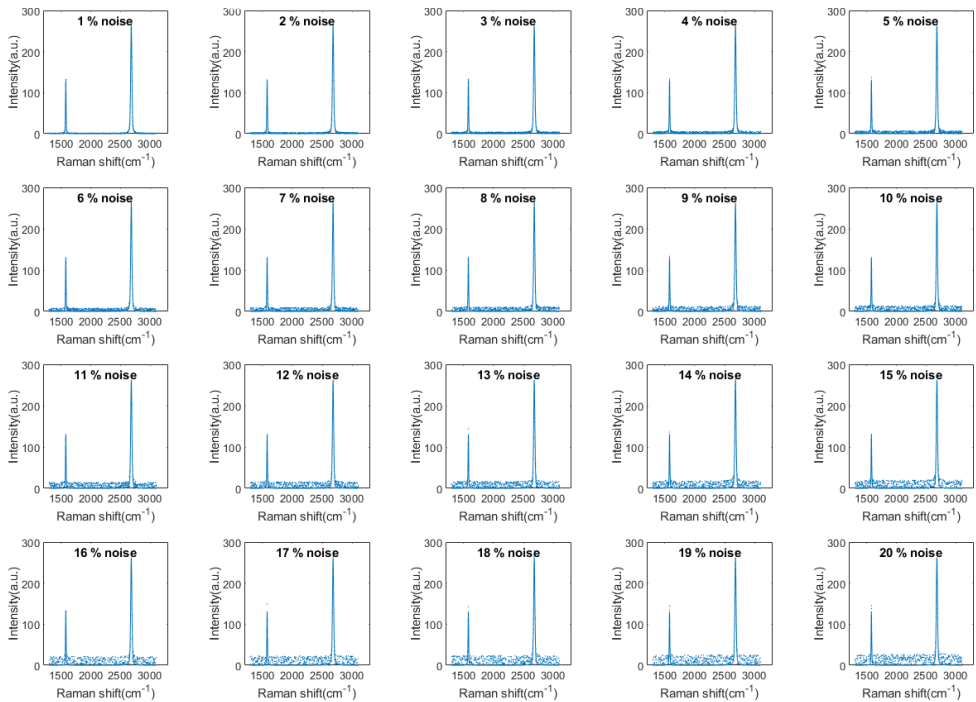

*Figure S2 Raman spectra with different noise levels from 1 % to 20 %.*

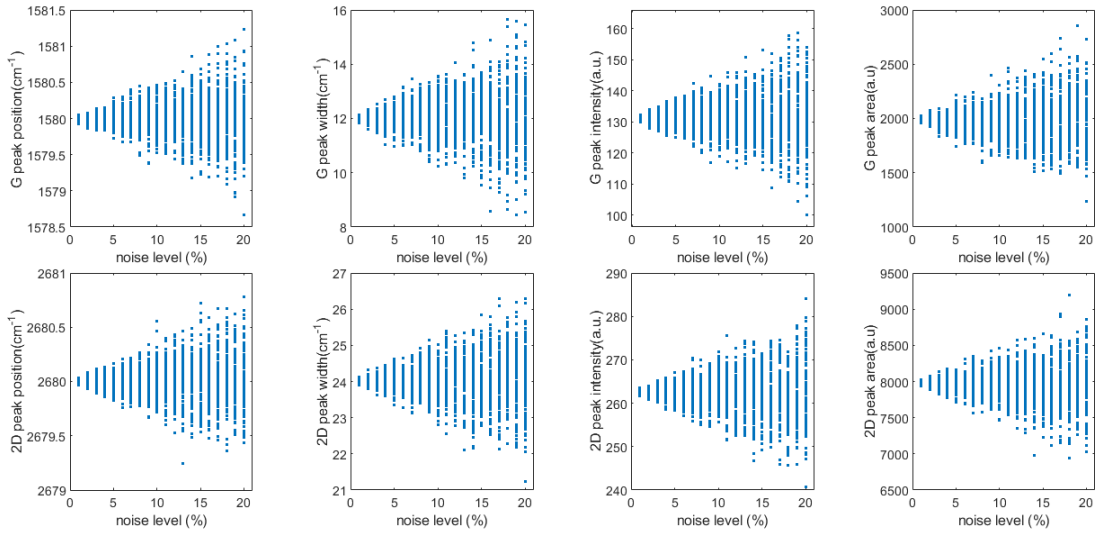

*Figure S3 The effect on noise level on the curve fitting results. The top row is the variation of G peak position, G peak FWHM, G peak intensity, G peak area, respectively. The bottom row is the variation of 2D peak position, 2D peak FWHM, 2D peak intensity, 2D peak area, respectively. At each noise level, we artificially generate 100 spectra and the spectra are fitted with Voigt profile. The results show that the extracted parameters are highly dependent on the noise levels of the collected spectra.*

*Table 1 Charge density and correlated parameters extracted from transport measurement.*

|    | CNP (V) | $n_0$ ($\times 10^{11} cm^{-2}$) | $\Delta n$ ($\times 10^{11} cm^{-2}$) | Mobility ($cm^2/Vs$) | Charge range : $n0 \pm \Delta n$) ($\times 10^{11} cm^{-2}$) |
|----|---------|----------------------------------|---------------------------------------|----------------------|--------------------------------------------------------------|
| C4 | 1.5     | 3.6                              | 2.2                                   | 17k                  | 1.4 to 5.8                                                   |
| C3 | 4.5     | 10.9                             | 7.2                                   | 5.2k                 | 3.7 to 18.1                                                  |
| C2 | 9.25    | 22.4                             | 5.7                                   | 2k                   | 16.7 to 28.1                                                 |
| C1 | ---     | ---                              | ---                                   | ---                  | 24 to 43                                                     |

To study graphene with different charge densities, we fabricated a graphene field-effect transistor (GFET) on a 90 nm SiO$_2$ substrate. The charge neutrality point (CNP), average accidental doping ($n_0$), charge fluctuation ($\Delta$n), mobility, and charge range are shown in Table 1. C1 is graphene on SiO2, which has been studied extensively and the charge doping level ranges from 24 to 43 $\times 10^{11}$ cm$^{-2}$. C2, C3, and C4 are from graphene fabricated on OTMS treated substrate.

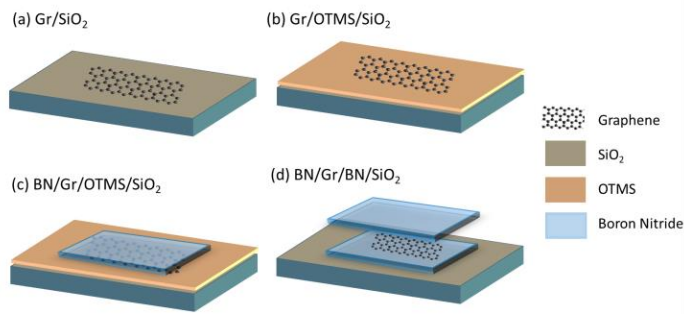

*Figure S4 Schematic of graphene in different dielectric environments. (a) graphene on SiO2 substrate. (b) graphene on OTMS-treated SiO2 substrate. (c) Boron nitride/Graphene on OTMS treated SiO2 substrate. (d) graphene encapsulated between boron nitride.*

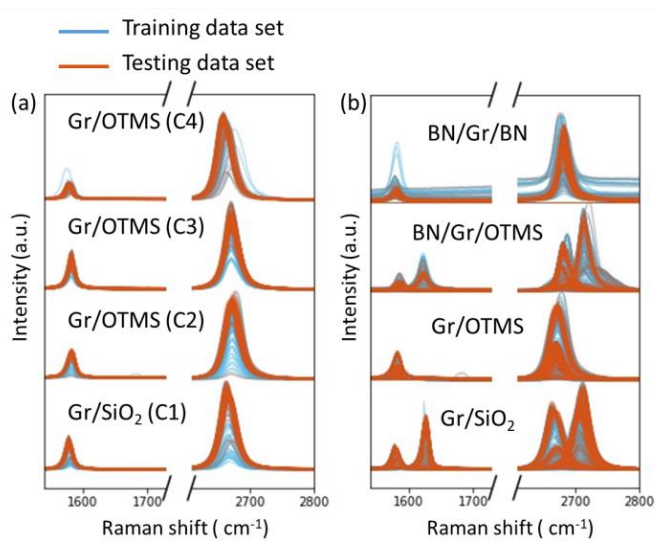

*Figure S5 Raman spectra plots of different classes. (a) 4 different classes of Raman spectra in the graphene charge variation dataset (GCV dataset). (b) 4 different classes of Raman spectra in the graphene dielectric variation dataset (GCV dataset). The Raman spectra of each class are gathered from different graphene samples. The blue lines are the training set and the red lines are the testing set.*

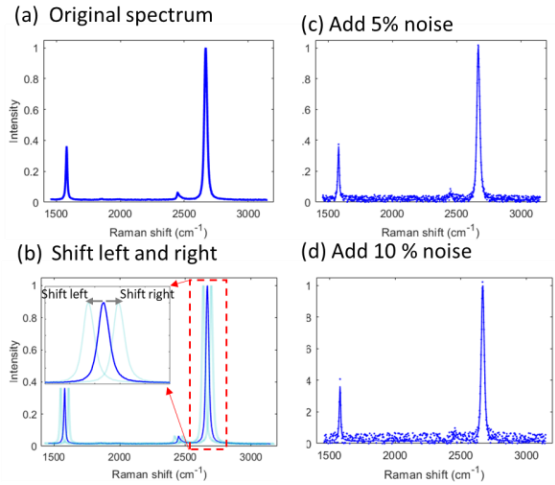

*Figure S 6 Illustration of data augmentation. (a) A representative Raman spectrum of graphene. (b) Left shift and right shift of the Raman peak. (c-d) add 5% and 10 % noise level to the original spectrum. The noise level is defined as the percentage of the noise fluctuation to the maximum intensity of the 2D peak in the Raman spectrum. The noise is generated by a random function that uniformly generates a random number up to the specified noise level.*

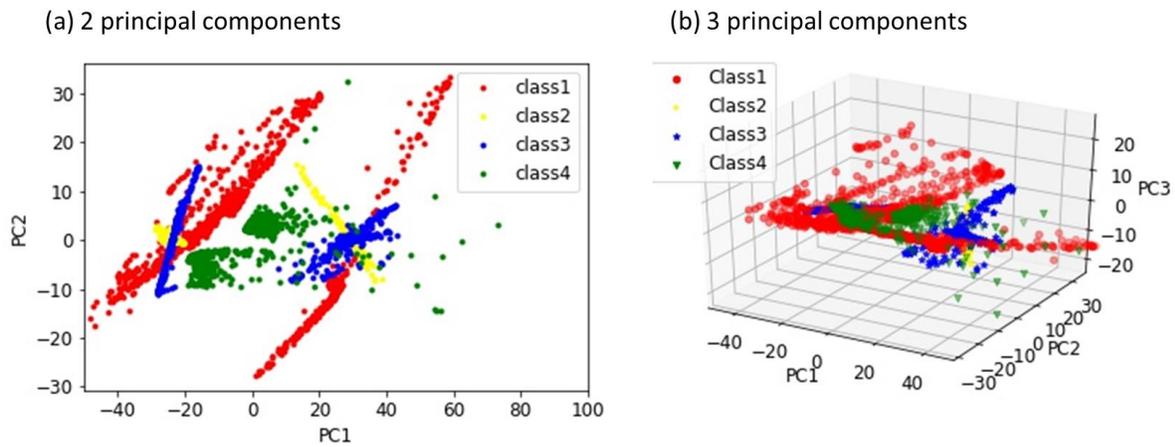

*Figure S7 Principal component analysis of the Raman spectra data. (a) Two pricipal component. (b) Three principal component.*

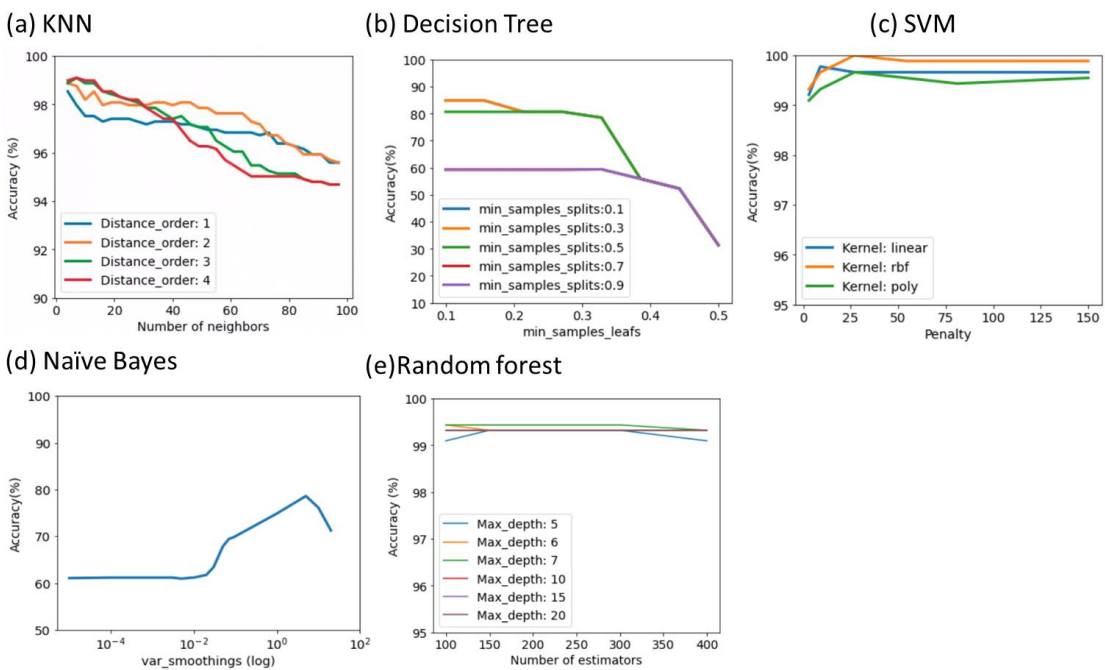

*Figure S8 Machine learning algorithms hyperparameter tuning. (a) KNN, the number of neighbors and the distance order vs. classification accuracy. (b) Decision tree, the minimum leaf size and and minimum split samples vs. accuracy. (c) SVM, the penalty and Kernal type vs. accuracy. (d) Naïve Bayes, the distribution variance parameter vs. accuracy. (e) Random forest, the number of estimators and maximum tree depth vs. accuracy.*

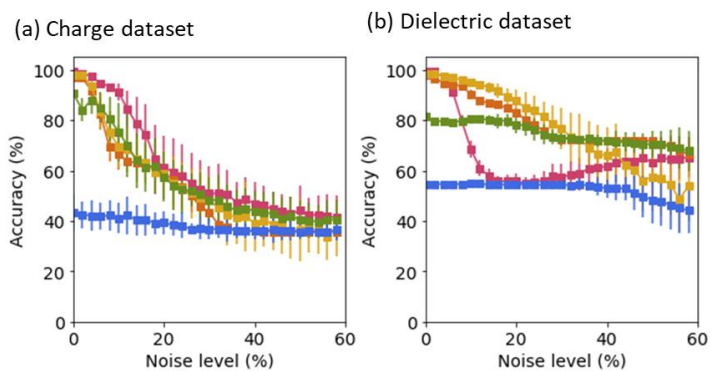

*Figure S9 (a) Testing performance as a function of noise on the charge dataset. (b) Testing performance as a function of noise on the dielectric dataset. Note that except adding noise, peak shift of spectra was also applied to data augmentation.*

## Model Summaries

Throughout this project, we experimented with four different deep network architectures CNN, FullCNN, MultiHead CNN, and Fully Connected neural network, all of which are detailed below.

## Fully Connected

This network is by far the simplest of the four architectures. It is 7 back-to-back linear layers. Each of the layers is followed by a relu activation and an appropriately sized batchnorm layer. The final output is then passed through a softmax function.

Table 2 Architecture summary of Fully connected neural network.

| Fully Connected | |
|---|---|
| Layer Type | Filter Size |
| FC | 1024 |
| FC | 512 |
| FC | 256 |
| FC | 128 |
| FC | 64 |
| FC | 16 |
| FC | 4 |

## CNN

This model uses 5 one dimensional convolutions after each other. The output is then flattened and passed to 2 fully connected layers. The exact details of the layers can be seen in the figure below. It is important to note that each of the 1D Conv layers is followed by a relu activation, batchnorm of the appropriate size, and then a 1D average pooling of size 2. Similarly, the first fully connected layer is also followed by a relu activation and a batchnorm. Finally, the last fully connected layer is followed by a relu activation and a softmax.

Table 3 Architecture summary of CNN

| CNN | | |
|---|---|---|
| Layer Type | Kernel Shape | Filter Count |
| 1D Conv | 9x1 | 2 |
| 1D Conv | 7x1 | 2 |
| 1D Conv | 7x1 | 4 |
| 1D Conv | 5x1 | 8 |
| 1D Conv | 3x1 | 12 |
| Flatten | - | - |
| FC | 128 | 1 |
| FC | 4 | 1 |

## FullCNN

This model is very closely related to the CNN model described above. It also starts with 5 1D convolution layers exactly as detailed before. However, after the fifth convolution, the output is then passed to a 1x1 convolution layer with a filter count of four. The output is this layer is then averaged across each filter yielding a single number for each filter, this is known as global average pooling. Once again, each convolution is followed by a relu activation, a batchnorm, and a 1d average pooling of size 2, and the last layer is also followed by a softmax function.

*Table 4 Architecture summary of Full CNN.*

| Full CNN | | |
| --- | --- | --- |
| Layer Type | Kernel Shape | Filter Count |
| 1D Conv | 9x1 | 2 |
| 1D Conv | 7x1 | 2 |
| 1D Conv | 7x1 | 4 |
| 1D Conv | 5x1 | 8 |
| 1D Conv | 3x1 | 12 |
| 1D Conv | 1x1 | 4 |
| Pooling | 1x19 | 1 |

**Multihead CNN**

This model is another variant of the CNN architecture. It passes the input into 3 different paths simultaneously, each of which containing its own, differently shaped, pair of 1d convolutional layers. All three outputs are then flattened and concatenated before being passed into the final fully connected layer. Each convolutional layer is followed by a relu activation, a batch normalization, and 1d max-pooling of size 2. The final layer is also followed by a softmax.

*Table 5 Architecture summary of MultiHead CNN.*

| MultiHead CNN | | | | | | | | |
| --- | --- | --- | --- | --- | --- | --- | --- | --- |
| Layer | Kernel Shape | Filter Count | Layer | Kernel Shape | Filter Count | Layer | Kernel Shape | Filter Count |
| 1D Conv | 3x1 | 16 | 1D Conv | 5x1 | 16 | 1D Conv | 7x1 | 16 |
| 1D Conv | 3x1 | 4 | 1D Conv | 5x1 | 4 | 1D Conv | 7x1 | 4 |
| Flatten | - | - | Flatten | - | - | Flatten | - | - |

| Layer | Size |
| --- | --- |
| FC | 4 |